\documentclass[11pt,a4paper]{article}
\usepackage{acl2015}
\usepackage{times}
\usepackage{url}
\usepackage{latexsym}
\usepackage{amsmath,graphicx,amsfonts,epsfig,epstopdf,datetime}
\usepackage{multirow}
\usepackage{url}
\usepackage{graphicx,epstopdf}
\usepackage{latexsym}
\usepackage[tight]{subfigure}
\usepackage{algorithm}
\usepackage{algorithmic}
\usepackage{amsmath} 
\usepackage{amssymb}  
\usepackage{cases}
\usepackage{float}
\usepackage{array}
\usepackage{mathrsfs}

\newcolumntype{C}[1]{>{\centering}p{#1}}

\DeclareMathOperator*{\argmax}{arg~max}
\DeclareMathOperator*{\argmin}{arg~min}

\title{Probabilistic Belief Embedding for Knowledge Base Completion}

\author{
Miao Fan$^{\dagger,\diamond,*}$, Qiang Zhou$^{\dagger}$, Andrew Abel$^{\star}$, Thomas Fang Zheng$^{\dagger}$ and Ralph Grishman$^{\diamond}$\\
$^{\dagger}$ CSLT, Division of Technical Innovation and Development,\\ Tsinghua National Laboratory for Information Science and Technology,\\ Tsinghua University, Beijing, 100084, China.\\
$^{\star}$ Computing Science and Mathematics School of Natural Science, \\University of Stirling, U.K.\\
$^{\diamond}$Proteus Group, New York University, NY, 10003, U.S.A.\\
 {\tt $^*$fanmiao.cslt.thu@gmail.com}
}
\date{}

\begin{document}
\maketitle
\begin{abstract}
 This paper contributes a novel embedding model which measures the probability of each candidate belief $\langle h,r,t,m\rangle$ in a large-scale knowledge repository via simultaneously learning distributed representations for entities ($h$ and $t$), relations ($r$), and even the words in relation mentions ($m$). It facilitates knowledge completion by means of simple vector operations to discover new beliefs. Given an imperfect belief, we can not only infer the missing entities, predict the unknown relations, but also tell the plausibility of that belief, just by exploiting the learnt embeddings of available evidence. To demonstrate the scalability and the effectiveness of our model, we conduct experiments on several large-scale repositories which contain hundreds of thousands of beliefs from WordNet, Freebase and NELL, and compare the results of a number of tasks, {\it entity inference}, {\it relation prediction} and {\it triplet classification}, with cutting-edge approaches. Extensive experimental results show that the proposed model outperforms other state-of-the-art methods, with significant improvements identified.
\end{abstract}

\section{Introduction}
\label{sec:intro}
Information extraction \cite{sarawagi2008information,Grishman:1997:IET:645856.669801}, the study of extracting structured beliefs from unstructured online texts to populate knowledge bases, has drawn much attention in recent years because of the explosive growth in the number of web pages. Thanks to long-term efforts of experts, crowd sourcing, and even machine learning techniques, several web-scale knowledge repositories, such as Wordnet\footnote{\url{http://wordnet.princeton.edu/}}, Freebase\footnote{\url{https://www.freebase.com/}} and NELL\footnote{\url{http://rtw.ml.cmu.edu/rtw/}}, have been constructed.  WordNet \cite{Miller1995} and Freebase \cite{Bollacker2007,Bollacker2008} follow the RDF format that represents each belief as a triplet, i.e. $\langle head~entity, relation, tail~entity\rangle$. NELL \cite{Carlson2010} extends each triplet with a $relation~mention$ which is a snatch of extracted free text to indicate the corresponding $relation$. Here we take a belief recorded in NELL as an example:
$\langle city:caroline,	citylocatedinstate,	stateorprovince:maryland, county~and~state~of\rangle$
, in which $county~and~state~of$ is the mention between the head entity $city:caroline$ and the tail entity $stateorprovince:maryland$, to indicate the relation $citylocatedinstate$. In some cases, NELL also provides the $confidence$ of each belief automatically learnt by machines.

Although colossal quantities of beliefs have been gathered, state-of-the-art work \cite{42024} reports that our knowledge bases are far from complete. For instance, nearly 97\% of people in the Freebase database have no records about their parents, whereas we can still find clues as to their immediate family in many cases by searching on the web and looking up their Wiki.

To populate the incomplete knowledge repositories, scientists either compare the performance of relation extraction between two named entities on manually annotated text datasets, such as  ACE\footnote{\url{http://www.itl.nist.gov/iad/mig/tests/ace/}} and MUC\footnote{\url{http://www.itl.nist.gov/iaui/894.02/related projects/muc/}}, or look for effective approaches for improving the accuracy of link prediction within the knowledge graphs constructed by the repositories without using extra free texts.

Recently, studies on text-based knowledge completion have benefited significantly from a paradigm known as Distantly Supervised Relation Extraction \cite{mintz2009distant} (DSRE), which bridges the gap between structured knowledge bases and unstructured free texts. It alleviates the labor of manual annotation by means of automatically aligning each triplet $\langle h,r,t\rangle$ from knowledge bases to the corresponding relation mention $m$ in free texts. However state-of-the-art research \cite{fan-EtAl:2014:P14-1} points out that DSRE still suffers from the problem of sparse and noisy features. Although Fan et al. \shortcite{fan-EtAl:2014:P14-1} fix the issue to some extent by making use of low-dimensional matrix factorization, their approach was identified as unable to handle large-scale datasets.

Fortunately, knowledge embedding techniques \cite{Bordes2011,Bordes2014} enable us to encode the high-dimensional sparse features into low-dimensional distributed representations. A simple but effective model is TransE \cite{Bordes2013a}, which trains a vector representation for each entity and relation in large-scale knowledge bases without considering any text information. Even though Weston et al. \cite{weston-EtAl:2013:EMNLP}, Wang et al. \cite{D14-1167} and Fan et al. \cite{fan2015jointly} broaden this field by adding word embeddings, there is still no comprehensive and elegant model that can integrate such large-scale heterogeneous resources to satisfy multiple subtasks of knowledge completion including {\it entity inference}, {\it relation prediction} and {\it triplet classification}.

Therefore, the contribution of this paper is a proposed novel embedding model which measures the probability of each belief $\langle h,r,t,m\rangle$ in large-scale repositories.  It breaks through the limitation of heterogeneous data, and establishes the connection between structured knowledge graphs and unstructured free texts. The distributed representations for entities ($h$ and $t$), relations ($r$), as well as the words in relation mentions ($m$) are simultaneously learnt within the uniform framework of the probabilistic belief embedding (PBE) we propose. Knowledge completion is facilitated by means of simple vector operations to discover new beliefs. Given an imperfect belief, we can not only infer the missing entities, predict the unknown relations, but tell the plausibility of the belief as well, just by means of the learnt vector representations of available data. To prove the effectiveness and the scalability of PBE, we perform extensive experiments with multiple tasks, including {\it entity inference}, {\it relation prediction} and {\it triplet classification}, for knowledge completion, and evaluate both our model and other cutting-edge approaches with appropriate metrics on several large-scale datasets which contain hundreds of thousands of beliefs from WordNet, Freebase and NELL. Detailed comparison results demonstrate that the proposed model outperforms other state-of-the-art approaches with significant improvements identified.

\section{Related Work}

Embedding-based inference models usually design various scoring functions $f_r(h, t)$ to measure the plausibility of a triplet $\langle h, r, t \rangle$. The lower the dissimilarity of the scoring function $f_r(h, t)$ is, the higher the compatibility of the triplet will be.

{\it Unstructured} \cite{Bordes2013a} is a naive model which exploits the occurrence information of the head and the tail entities without considering the relation between them. It defines a scoring function $||{\bf h}-{\bf t}||$, and this model obviously can not discriminate between a pair of entities involving different relations. Therefore, {\it Unstructured} is commonly regarded as the baseline approach.

{\it Distance Model (SE)} \cite{Bordes2011} uses a pair of matrices $(W_{rh}, W_{rt})$, to characterize a relation $r$. The dissimilarity of a triplet is calculated by $||W_{rh}{\bf h} - W_{rt}{\bf t}||_1$. As identified by Socher et al. \cite{Socher2013}, the separating matrices $W_{rh}$ and $W_{rt}$ weaken the capability of capturing correlations between entities and corresponding relations, despite the model taking the relations into consideration.

{\it Single Layer Model}, proposed by Socher et al. \cite{Socher2013} thus aims to alleviate the shortcomings of the {\it Distance Model} by means of the non-linearity of a single layer neural network $g(W_{rh}{\bf h} + W_{rt}{\bf t} + {\bf b}_r)$, in which $g = tanh$. The linear output layer then gives the scoring function: ${\bf u}^T_rg(W_{rh}{\bf h} + W_{rt}{\bf t} + {\bf b}_r)$.

{\it Bilinear Model} \cite{Sutskever2009,Jenatton2012} is another model that tries to fix the issue of weak interaction between the head and tail entities caused by the {\it Distance Model} with a relation-specific bilinear form: $f_r(h, t) = {\bf h}^TW_r{\bf t}$.

{\it Neural Tensor Network (NTN)} \cite{Socher2013} designs a general scoring function:  $f_r(h, t) = {\bf u}^T_rg({\bf h}^TW_r{\bf t}+ W_{rh}{\bf h} + W_{rt}{\bf t} + {\bf b}_r)$, which combines the {\it Single Layer} and {\it Bilinear} Models. This model is more expressive as the second-order correlations are also considered in the non-linear transformation function, but the computational complexity is rather high.

{\it TransE} \cite{Bordes2013a} is a canonical model different from all the other prior arts, which embeds relations into the same vector space as entities by regarding the relation $r$ as a translation from $h$ to $t$, i.e. ${\bf h} + {\bf r} = {\bf t}$. It works well on beliefs with the ONE-TO-ONE mapping property but performs badly with multi-mapping beliefs. Given a series of facts associated with a ONE-TO-MANY relation $r$, e.g. ${\langle h, r, t_1 \rangle, \langle h, r, t_2 \rangle, ..., \langle h, r, t_m \rangle}$, {\it TransE} tends to represent the embeddings of entities on the MANY-side as extremely close to each other with very little discrimination.

{\it TransM} \cite{fan2014transition} exploits the structure of the whole knowledge graph, and adjusts the learning rate, which is specific to each relation based on the multiple mapping property of the relation.

{\it TransH} \cite{DBLP:conf/aaai/WangZFC14} is, to the best knowledge of the authors, the state of the art approach. It improves {\it TransE} by modeling a relation as a hyperplane, which makes it more flexible with regard to modeling beliefs with multi-mapping properties.

Due to the diverse feature spaces between unstructured texts and structured beliefs, one key challenge of connecting natural language and knowledge is to project the features into the same space and to merge them together for knowledge completion. Fan et al. \cite{fan2015jointly} have recently proposed {\it JRME} to jointly learn the embedding representations for both relations and mentions in order to predict unknown relations between entities in NELL. However, the functionality of their latest method is limited to the relation prediction task (see section \ref{sec:exp-relpred}), as the correlations between entities and relations are ignored. Therefore, we desire a comprehensive model that can simultaneously consider entities, relations and even the relation mentions, and can integrate the heterogeneous resources to support multiple subtasks of knowledge completion, such as {\it entity inference}, {\it relation prediction} and {\it triplet classification}.

\begin{figure*}
\centering
\includegraphics[width=0.8\textwidth]{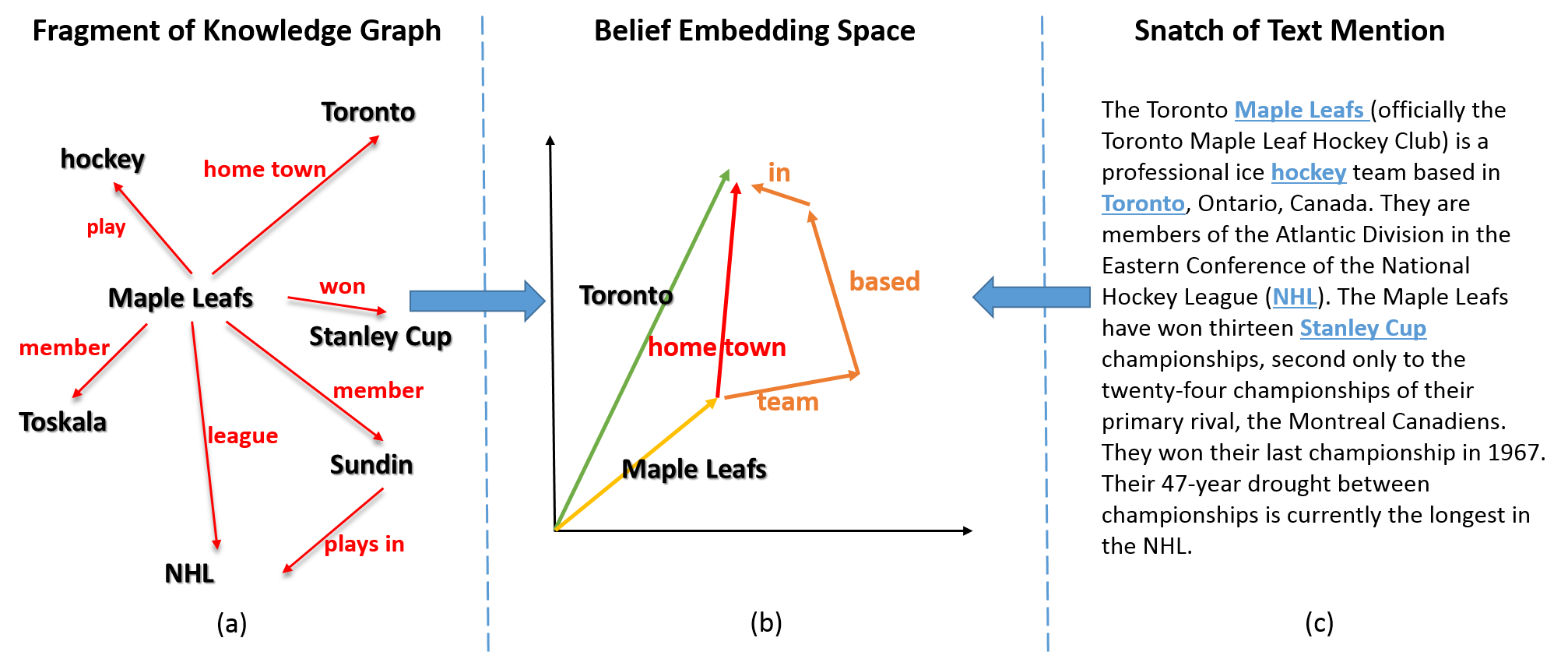}\\
\caption{The whole framework of belief embedding. (a) shows a fragment of knowledge graph; (c) is a snatch of Wiki which describes the knowledge graph of (a); (b) illustrates how the belief $\langle Maple~Leafs, home~town, Toronto, team~based~in\rangle$ is projected into the same embedding space.}
\end{figure*}

\section{Model}
The intuition of the subsequent theory is that: Not each belief we have learnt, i.e. $\langle head~entity, relation, tail~entity, mention\rangle$ abbreviated as $\langle h, r, t, m\rangle$, is perfect and complete enough \cite{fan2015learning}. We thus explore modeling the probability of each belief, i.e. $Pr(h,r,t,m)$. It is assumed that $Pr(h,r,t,m)$ is collaboratively influenced by $Pr(h|r,t)$, $Pr(t|h,r)$ and $Pr(r|h,t,m)$, where $Pr(h|r,t)$ stands for the conditional probability of inferring the head entity $h$ given the relation $r$ and the tail entity $t$, $Pr(t|h,r)$ represents the conditional probability of inferring the tail entity $t$ given the head entity $h$ and the relation $r$, and $Pr(r|h,t,m)$ denotes the conditional probability of predicting the relation $r$ between the head entity $h$ and the tail entity $t$ with the relation mention $m$ extracted from free texts. Therefore, we define that the probability of a belief equals to the geometric mean of $Pr(h|r,t)$$Pr(r|h,t,m)$$Pr(t|h,r)$ as shown in the subsequent equation,
\begin{equation}
Pr(h,r,t,m) = \sqrt[3]{Pr(h|r,t)Pr(r|h,t,m)Pr(t|h,r)}.
\end{equation}

We assume that we have a certain repository $\Delta$, such as WordNet, which contains thousands of beliefs validated by experts. The learning object is intuitively set to maximize $\mathcal{L}_{max}$, where
\begin{equation}
\mathcal{L}_{max} = \prod_{\langle h,r,t,m\rangle \in \Delta} Pr(h,r,t,m).
\end{equation}
In most cases, we can automatically build much larger but imperfect knowledge bases as well via crowdsouring (Freebase) and machine learning techniques (NELL). However, each belief of NELL has a confidence-weighted score $c$ to indicate its plausibility to some extent. Therefore, we propose an alternative goal which aims to minimize $\mathcal{L}_{min}$, in which
\begin{equation}
\mathcal{L}_{min} = \prod_{\langle h,r,t,m,c\rangle \in \Delta} \frac{1}{2}[ Pr(h,r,t,m) - c]^2.
\end{equation}

To facilitate the optimization progress, we prefer using the log likelihood of $\mathcal{L}_{max}$ and $\mathcal{L}_{min}$, and the learning targets can be further processed as follows,
\begin{equation}
\begin{split}
&~~~~~\argmax_{h,r,t,m} ~~~~ \log \mathcal{L}_{max} \\
&= \argmax_{h,r,t,m} \sum_{\langle h,r,t,m \rangle \in \Delta}\log Pr(h,r,t,m)\\
&= \argmax_{h,r,t,m} \sum_{\langle h,r,t,m\rangle \in \Delta} \frac{1}{3} ~[\log Pr(h|r,t) \\
& ~~~~~+ \log Pr(r|h,t,m) + \log Pr(t|h,r)];\\
\end{split}
\end{equation}

\begin{equation}
\begin{split}
&~~~~~\argmin_{h,r,t,m} ~~~~\log \mathcal{L}_{min} \\
&= \argmin_{h,r,t,m} \sum_{\langle h,r,t,m,c \rangle \in \Delta} \frac{1}{2}[\log Pr(h,r,t,m) - \log c]^2\\
&= \argmin_{h,r,t,m}\sum_{\langle h,r,t,m,c\rangle \in \Delta} \frac{1}{2}\{\frac{1}{3} ~[ \log Pr(h|r,t)\\
&~~~~~ + \log Pr(r|h,t,m) + \log Pr(t|h,r)] - \log c\}^2.\\
\end{split}
\end{equation}
The advantage of the conversions above is that we can separate the factors out, compared with Equation (1).  Therefore, the remaining challenge is to identify the approaches to use to model $Pr(h|r,t)$, $Pr(r|h,t,m)$ and $Pr(t|h,r)$.

$Pr(r|h,t,m)$ leverages the evidences from two different resources to predict the relation. If the concurrence of the two entities ($h$ and $t$) in knowledge bases is independent of the appearance of the relation mention $m$ from free texts, we can factorize $Pr(r|h,t,m)$ as shown by Equation (6):
\begin{equation}
Pr(r|h,t,m) = Pr(r|h,t)Pr(r|m).
\end{equation}

We then need to consider formulating $Pr(h|r,t)$, $Pr(r|h,t)$, $Pr(t|h,r)$ and $Pr(r|m)$, respectively.

Figure 1(a) illustrates the traditional way of recording knowledge as triplets. The triplets $\langle h, r, t \rangle$ can construct a knowledge graph in which entities ($h$ and $t$) are nodes and the relation ($r$) between them is a directed edge from the head entity ($h$) to the tail entity ($t$). This kind of symbolic representation, whilst being very efficient for storing, is not flexible enough for statistical learning approaches \cite{Bordes2011}. However, once each of the elements, including entities and relations in the knowledge repository, are projected into the same embedding space, we can use:
\begin{equation}
\mathscr{D}(h, r, t) = - {\bf ||h + r - t||} + \alpha,
\end{equation}
which is a simple vector operation to measure the distance between ${\bf h + r}$ and ${\bf t}$, where $h$, $r$ and $t$ are encoded in $d$ dimensional vectors, and $\alpha$ is the bias parameter. To estimate the conditional probability of appearing $t$ given $h$ and $r$, i.e. $Pr(t|h,r)$, however, we need to adopt the softmax function as follows,
\begin{equation}
Pr(t|h,r) = \frac{\exp^{\mathscr{D}(h,r,t)}}{\sum_{t' \in E_t}{\exp^{\mathscr{D}(h,r,t')}}},
\end{equation}
where $E_t$ is the set of tail entities which contains all possible entities $t'$ appearing in the tail position. Similarly, we can regard $Pr(h|r,t)$ and $Pr(r|h,t)$ as
\begin{equation}
Pr(h|r,t) = \frac{\exp^{\mathscr{D}(h,r,t)}}{\sum_{h' \in E_h}{\exp^{\mathscr{D}(h',r,t)}}}
\end{equation}
and
\begin{equation}
Pr(r|h,t) = \frac{\exp^{\mathscr{D}(h,r,t)}}{\sum_{r' \in R}{\exp^{\mathscr{D}(h,r',t)}}},
\end{equation}
in which $E_h$ is the set of head entities which contains all possible entities $h'$ appearing in the head position, and  $R$ is the set of all candidate relations $r'$.

In addition to this, Figure 1(c) shows that free texts can provide fruitful contexts between two recognized entities, but the one-hot\footnote{\url{http://en.wikipedia.org/wiki/One-hot}} feature space is rather high and sparse. Therefore, we can also project each words in relation 'mentions' into the same embedding space of entities and relations. To measure the similarity between the mention $m$ and the corresponding relation $r$, we adopt the inner product of their embeddings as shown by Equation (11),
\begin{equation}
\mathscr{F}(r,m) = {\bf W}^T{\bf \phi}(m){\bf r} + \beta,
\end{equation}
where ${\bf W}$ is the matrix of $\mathbb{R}^{{n_v} \times d}$ containing $n_v$ vocabularies with $d$ dimensional embeddings, ${\bf \phi}(m)$ is the sparse one-hot representation of the mention indicating the absence or presence of words, $r \in \mathbb{R} ^ d$ is the embedding of relation $r$, and $\beta$ is the bias parameter. Similar to Equations (8), (9) and (10), the conditional probability of predicting relation $r$ given mention $m$, i.e. $Pr(r|m)$ can be defined as,
\begin{equation}
Pr(r|m) = \frac{\exp^{\mathscr{F}(r,m)}}{\sum_{r' \in R}{\exp^{\mathscr{F}(r',m)}}}.
\end{equation}

Overall, this section has shown that we can model the probability of a belief via jointly embedding the entities, relations and even the words in mentions, as demonstrated by Figure 1(b).
\begin{table*}
\centering

\begin{tabular}{|c|c|c|c|}
  \hline
  {\bf DATASET} & {\bf NELL-50K} & {\bf WN-100K} & {\bf FB-500K}\\
  \hline
  \hline
   \#(ENTITY) & 29,904  & 38,696  & 14,951\\
  \#(RELATION) & 233  & 11  & 1,345\\
  \#(VOCABULARY) & 8,948 & - & - \\
  \#(TRAINING EX.) & 57,356  & 112,581  & 483,142\\
  \#(VALIDATING EX.) & 10,710 & 5,218  & 50,000\\
  \#(TESTING EX.) & 10,711 & 21,088 & 59,071\\
  \#(TC VALIDATING EX.) & 21,420 & 10,436  & 100,000\\
  \#(TC TESTING EX.) & 21,412 & 42,176  & 118,142\\
  \hline
\end{tabular}
\caption{Statistics of the datasets used for the subtasks, i.e. {\it entity inference}, {\it relation prediction} and {\it triplet classification} of knowledge completion.}
\end{table*}

\section{Algorithm}
To search for the optimal solutions of Equation (4) and (5), we can use {\it Stochastic Gradient Descent} (SGD) to update the embeddings of entities, relations and words of mentions in iterative fashion. However, it is computationally intensive to calculate the normalization terms in $Pr(h|r,t)$, $Pr(r|h,t)$, $Pr(t|h,r)$ and $Pr(r|m)$ according to the definitions made by Equation (8), (9), (10) and (12) respectively. For instance, if we directly calculate the value of $Pr(h|r,t)$ for just one belief, tens of thousands $\exp^{\mathscr{D}(h',r,t)}$ need to be re-valued, as there are tens of thousands of candidate entities $h'$ in $E_h$. Inspired by the work of Mikolov et al. \cite{mikolov2013distributed}, we have developed an efficient approach that adopts the negative sampling technique to approximate the conditional probability functions, i.e. Equations (8), (9), (10) and (12), by being transformed to binary classification problems as shown respectively by the subsequent equations,
\begin{equation}
\begin{split}
&\log Pr(h|r,t) \approx \log Pr(1|h,r,t) \\
& + \sum^k_{i = 1}{\mathbb{E}_{h'_i Pr(h' \in E_h)}\log Pr(0|h'_i, r, t)},\\
\end{split}
\end{equation}
\begin{equation}
\begin{split}
&\log Pr(t|h,r) \approx \log Pr(1|h,r,t) \\
& + \sum^k_{i = 1}{\mathbb{E}_{t'_i Pr(t' \in E_t)}\log Pr(0|h, r, t'_i)},\\
\end{split}
\end{equation}
\begin{equation}
\begin{split}
&\log Pr(r|h,t) \approx \log Pr(1|h,r,t) \\
&+ \sum^k_{i = 1}{\mathbb{E}_{r'_i Pr(r' \in R)}\log Pr(0|h, r'_i, t)},\\
\end{split}
\end{equation}
\begin{equation}
\begin{split}
&\log Pr(r|m) \approx \log Pr(1|r,m) \\
&+ \sum^k_{i = 1}{\mathbb{E}_{r'_i Pr(r' \in R)}\log Pr(0|r'_i, m)},\\
\end{split}
\end{equation}
where we sample $k$ negative beliefs and discriminate them from the positive case. For the simple binary classification problems mentioned above, we choose the logistic function with the offset $\epsilon$ shown in Equation (17) to estimate the probability that the given triplet $\langle h, r, t\rangle$ is correct:
\begin{equation}
Pr(1|h,r,t) = \frac{1}{1 + \exp^{-\mathscr{D}(h,r,t)}} + \epsilon,
\end{equation}
and with the offset $\eta$ shown in Equation (18) to tell the probability of the occurrence of $r$ and $m$:
\begin{equation}
Pr(1|r,m) = \frac{1}{1 + \exp^{-\mathscr{F}(r,m)}} + \eta.
\end{equation}

\section{Experiment}
Besides its access to the efficient SGD algorithm, the learnt embeddings by {\it PBE} can contribute more effectiveness on multiple subtasks of knowledge completion, such as entity inference, relation prediction, and triplet classification.

\subsection{Dataset}
To demonstrate the wide adaptability and significant effectiveness of our approach, we prepare three datasets as shown by Table 1, i.e. {\bf NELL-50K}, {\bf WN-100K}, {\bf FB-500K} from the repositories of NELL \cite{carlson-aaai}, WordNet \cite{Miller1995} and Freebase \cite{Bollacker2007,Bollacker2008} respectively.
The NELL \cite{NELL-aaai15} designed and maintained by Carnegie Mellon University is an outstanding system which runs 24 hours/day and never stops learning the beliefs on the Web. We use a relatively small dataset {\bf NELL-50K} which contains about 50 thousand confidence-weighted beliefs from NELL. Each belief of {\bf NELL-50K} has a relation mention $m$ in addition to a triplet $\langle h, r, t \rangle$. {\bf WN-100K} is made by experts, and owns only 11 kinds of relations but much more entities. Therefore, it is a sparse repository in which fewer entities have connections. The third dataset ({\bf FB-500K}\footnote{We change the original name of the dataset ({\bf FB15K}), so as to follow the naming conventions in our paper. Related studies on this dataset can be looked up from the website \url {https://www.hds.utc.fr/everest/doku.php?id=en:transe}}) we adopt is released by Bordes et al. \cite{Bordes2013a}. It is a large crowdsourcing dataset extracted from Freebase, in which each belief is a triplet without a confidence score.

As the words in relation mentions will be further concerned in the {\it relation prediction} subtask, we also show the vocabulary size of each dataset. However, {\bf WN-100K} and {\bf FB-500K} only contain triplets as beliefs, so their vocabulary sizes are null.

For the {\it triplet classification} subtask, the head or the tail entity can be randomly replaced with another one to produce a negative training example. But in order to build much tough validation and testing datasets, we constrain that the picked entity should once appear at the same position. For example, {\it (Pablo Picaso, nationality, U.S.)} is a potential negative example rather than the obvious nonsense {\it (Pablo Picaso, nationality, Van Gogh)}, given a positive triplet {\it (Pablo Picaso, nationality, Spain)}.

\begin{table*}
  \centering

  \begin{tabular}{*{5}{|c|}}
    \hline
    {\bf DATASET} & \multicolumn{4}{|c|}{\bf NELL-50K}\\
    \hline
    \hline
    {\multirow{2}*{\bf METRIC}}
    &  \multicolumn{2}{|c|}{\em MEAN RANK} & \multicolumn{2}{|c|}{\em MEAN HIT@10}\\
   &  {\em Raw} & {\em Filter} & {\em Raw} & {\em Filter}\\
   \hline
   \hline
   {\bf TransE}  & 2,436 / 29,904 & 2,426 / 29,904 & 18.9\% & 19.6\%  \\
   {\bf TransM}  & 2,296 / 29,904 & 2,285 / 29,904 & 20.5\% & 21.3\%  \\
   {\bf TransH}  & 2,185 / 29,904& 2,072 / 29,904 & 21.6\% & {\bf 28.8\%}  \\
   \hline
   {\bf PBE}   & {\bf 2,078} / 29,904 & {\bf 1,996} / 29,904 & {\bf 22.5\%} & 26.4\% \\
   \hline
  \end{tabular}
    \caption{Entity inference results on the {\bf NELL-50K} dataset. We compared our proposed PBE with the state-of-the-art method TransH and other prior arts mentioned in Section 2.}

\end{table*}

\begin{table*}
  \centering

  \begin{tabular}{*{5}{|c|}}
    \hline
    {\bf DATASET} & \multicolumn{4}{|c|}{\bf WN-100K }\\
    \hline
    \hline
    {\multirow{2}*{\bf METRIC}}
    &  \multicolumn{2}{|c|}{\em MEAN RANK} & \multicolumn{2}{|c|}{\em MEAN HIT@10}\\
   &  {\em Raw} & {\em Filter} & {\em Raw} & {\em Filter}\\
   \hline

   {\bf TransE}  & 10,623 /  38,696 & 10,575 /   38,696 & 3.8\% & 4.1\% \\ 
   {\bf TransM}  & 14,586 /  38,696 & 13,276 /   38,696 & 1.8\% & 2.0\% \\ 
   {\bf TransH}  & 12,542/  38,696 & 12,463 /   38,696 & 2.3\% &  2.6\%  \\
   \hline
   \hline

   {\bf PBE}   & {\bf 8,462} / 38,696 & {\bf 8,409} / 38,696 & {\bf 9.0\%} &  {\bf 10.1}\% \\
   \hline
  \end{tabular}
      \caption{Entity inference results on the {\bf WN-100K} dataset. We compared our proposed PBE with the state-of-the-art method TransH and other prior arts mentioned in Section 2.}

\end{table*}

\begin{table*}
  \centering

  \begin{tabular}{*{5}{|c|}}
    \hline
    {\bf DATASET} & \multicolumn{4}{|c|}{\bf FB-500K}\\
    \hline
    \hline
    {\multirow{2}*{\bf METRIC}}
    &  \multicolumn{2}{|c|}{\em MEAN RANK} & \multicolumn{2}{|c|}{\em MEAN HIT@10}\\
   &  {\em Raw} & {\em Filter} & {\em Raw} & {\em Filter}\\
   \hline
   \hline
   {\bf TransE}  & 243 / 14,951& 125 / 14,951& 34.9\% & 47.1\%  \\
   {\bf TransM}  & 196 / 14,951 & 93 / 14,951& 44.6\% & 55.2\%  \\
   {\bf TransH}  & 211 / 14,951& 84 / 14,951& 42.5\% & 58.5\%  \\
   \hline
   {\bf PBE}   & {\bf 165} / 14,951 & {\bf 61} / 14,951& {\bf 50.5\%} & {\bf 64.6\%} \\
   \hline
  \end{tabular}
     \caption{Entity inference results on the {\bf FB-500K} dataset. We compared our proposed PBE with the state-of-the-art method TransH and other prior arts mentioned in Section 2.}
\end{table*}

\begin{table*}
\centering
\begin{tabular}{|c|c|c|c|c|c|c|c||c|c|c|}
  \hline
   {\bf DATASET} & \multicolumn{3}{|c|}{\bf NELL-50K} & \multicolumn{3}{|c|}{\bf WN-100K } & \multicolumn{3}{|c|}{\bf FB-500K }\\
   \hline
   \hline
  {\bf METRIC} & {\it AVG. R.} & {\it HIT@10} & {\it HIT@1}  & {\it AVG. R.} & {\it HIT@10} & {\it HIT@1} & {\it AVG. R.} & {\it HIT@10} & {\it HIT@1}\\
  \hline
  \hline
   {\bf TransE} & 131.8 & 16.3\% & 3.0\% & 3.8 & 98.3\% & 15.1\%& 762.7 & 7.3\% & 1.9\%\\
   {\bf TransM}   & 70.2 & 18.9\% & 4.3\% & 4.6 & 97.5\% & 14.8\% & 402.3 & 13.4\% & 3.2\% \\
   {\bf TransH}   & 46.3 & 20.0\% & 5.1\% & 3.4 & 99.0\% & 19.3\% & 79.5 & 39.2\% & 15.6\%\\
   {\bf JRME}  & 6.2 & 87.8\%  & 60.2\% & 3.9 & 99.0\% & 15.9\% & 60.9 & 27.4\% & 7.2\%\\
  \hline
   {\bf PBE} & {\bf 2.5}& {\bf 96.6\%}&{\bf 78.3\%} & {\bf 2.0} & {\bf 99.1\%} & {\bf 72.6\%} & {\bf 2.6} & {\bf 97.3\%} & {\bf 66.7\%}\\
   \hline
\end{tabular}
\caption{Performance of relation prediction on TransE, TransM, TransH, JRME and PBE evaluated by the metrics of Average Rank, Hit@10 and Hit@1 with {\bf NELL-50K}, {\bf WN-100K}, and {\bf FB-500K} dataset.}
\end{table*}
\begin{table*}
\centering
\begin{tabular}{|c|c|c|c|}
  \hline
  {\bf DATASET} & {\bf NELL-50K} & {\bf WN-100K} & {\bf FB-500K}\\
  \hline
  \hline
  {\bf METRIC} & {\it ACC.} & {\it ACC.}  & {\it ACC.} \\
  \hline
  \hline
  {\bf TransE} & 80.5\%  & 64.2\% & 79.9\%\\
  {\bf TransM} & 82.0\%  & 57.2\% & 85.8\%\\
  {\bf TransH} & 83.6\%  & 59.5\% & 87.7\%\\
  \hline
  {\bf PBE} & {\bf 90.2\%} & {\bf 67.8\%} & {\bf 92.6\%}\\
  \hline
\end{tabular}
  \caption{The accuracy of triplet classification compared with several latest approaches: TransH, TransM and TransE, with {\bf NELL-50K}, {\bf WN-100K}, and {\bf FB-500K} dataset.}
\end{table*}

\subsection{Entity inference}
One of the benefits of knowledge embedding is that simple vector operations can apply to entity inference which contributes to knowledge graph completion. Given a wrecked triplet, like $\langle h, r, ? \rangle$ or $\langle ?, r, t \rangle$, the subtask needs to compute the $\argmax_{h \in E_h} Pr(h | r, t)$, with the help of the entity and relation embeddings. In the meanwhile, $\argmax_{t \in E_t} Pr(t|h, r)$ will help us to find the best tail entity given the head entity $h$ and the relation $r$.

\subsubsection{Metric}
For each testing belief, all the other entities that appear in the training set take turns to replace the head entity. Then we get a bunch of candidate triplets. The plausibility of each candidate triplet is firstly computed by various scoring functions, such as $Pr(h|r, t)$ in {\it PBE}, and then sorted in ascending order. Finally, we locate the ground-truth triplet and record its rank. This whole procedure runs in the same way when replacing the tail entity, so that we can gain the mean results. We use two metrics, i.e. {\it Mean Rank} and {\it Mean Hit@10} (the proportion of ground truth triplets that rank in Top 10), to measure the performance. However, the results measured by those metrics are relatively {\it raw}, as the procedure above tends to generate false negative triplets. In other words, some of the candidate triplets rank rather higher than the ground truth triplet just because they also appear in the training set. We thus filter out those triplets to report more reasonable results.
\subsubsection{Performance}
We compare {\it PBE} with the state-of-the-art {\it TransH} \cite{DBLP:conf/aaai/WangZFC14}, {\it TransM} \cite{fan2014transition}, {\it TransE} \cite{Bordes2013a} mentioned in Section 2 evaluated on {\bf NELL-50K}, {\bf WN-100K} and {\bf FB-500K} datasets. We tune the parameters of each previous model based on the validation set, and select the combination of parameters which leads to the best performance. Table 2, 3 and 4 demonstrate that {\it PBE} outperforms the prior arts on almost all the metrics. Overall, it achieves significant improvements (relative increment) on all three datasets, with
{\bf NELL-50K}: \{{\it Mean Rank Raw}: $4.9\%\Uparrow$, {\it Hit@10 Raw}: $4.2\%\Uparrow$, {\it Mean Rank Filter}: $3.7\%\Uparrow$, {\it Hit@10 Filter}: $8.3\%\Downarrow$\},
{\bf WN-100K}: \{{\it Mean Rank Raw}: $20.3\%\Uparrow$, {\it Hit@10 Raw}: $136.8\%\Uparrow$, {\it Mean Rank Filter}: $20.5\%\Uparrow$, {\it Hit@10 Filter}: $146.3\%\Uparrow$\}
and {\bf FB-500K}: \{{\it Mean Rank Raw}: $15.8\%\Uparrow$, {\it Hit@10 Raw}: $27.3\%\Uparrow$, {\it Mean Rank Filter}: $13.3\%\Uparrow$, {\it Hit@10 Filter}: $10.4\%\Uparrow$\}.

\subsection{Relation prediction}
\label{sec:exp-relpred}
The scenario of this subtask is that: given a pair of entities and the text mentions indicating the semantic relations between them, i.e. $\langle h, ?, t, m \rangle$, this subtask computes the $\argmax_{r \in R} Pr(r|h, t)Pr(r|m)$ to predict the best relations.

\subsubsection{Metric}
We compare the performances between our models and other state-of-the-art approaches, with the metrics as follows,

{\it Average Rank}: Each candidate relation will gain a score calculated by Equation (7). We sort them in ascent order and compare with the corresponding ground-truth belief. For each belief in the testing set, we get the rank of the correct relation. The average rank is an aggregative indicator, to some extent, to judge the overall performance on relation extraction of an approach.

{\it Hit@10}: Besides the average rank, scientists from the industrials concern more about the accuracy of extraction when selecting Top10 relations. This metric shows the proportion of beliefs that we predict the correct relation ranked in Top10.

{\it Hit@1}: It is a more strict metric that can be referred by automatic system, since it demonstrates the accuracy when just picking the first predicted relation in the sorted list.
\subsubsection{Performance}
Table 5 illustrates the results of experiments on relation prediction with all the three datasets, respectively. We find out that text mentions within the {\bf NELL-50K} contribute a lot on predicting the correct relations. All of results show that {\it PBE} performs best compared with the latest approaches. The relative increments are
{\bf NELL-50K}: \{{\it Mean Rank}: $59.7\%\Uparrow$, {\it Hit@10}: $10.0\%\Uparrow$, {\it Hit@1}: $30.0\%\Uparrow$\},
{\bf WN-100K}: \{ {\it Mean Rank}: $41.1\%\Uparrow$, {\it Hit@10}: $0.1\%\Uparrow$, {\it Hit@1}: $276.2\%\Uparrow$ \}
and {\bf FB-500K}: \{ {\it Mean Rank}: $95.7\%\Uparrow$, {\it Hit@10}: $148.2\%\Uparrow$, {\it Hit@1}: $327.6\% \Uparrow$ \}.

\subsection{Triplet classification}
Triplet classification is another inference related task proposed by Socher et al. \cite{Socher2013} which focuses on searching a relation-specific threshold $\sigma_r$ to identify whether a triplet $\langle h, r, t \rangle$ is plausible. If the probability of a testing triplet ($h, r, t$) computed by $Pr(h|r,t)Pr(r|h,t)Pr(t|h,r)$ is below the relation-specific threshold $\sigma_r$, it is predicted as positive, otherwise negative.
\subsubsection{Metric}
We use {\it classification accuracy} to measure the performances among the competing methods. Specifically, we sum up the correctness of each triplet $\langle h, r, t \rangle$ via comparing the probability of the triplet and the relation-specific threshold $\sigma_r$, which can be searched via maximizing the classification accuracy on the validation triplets which belong to the relation $r$.

\subsubsection{Performance}

Compared with several of the latest approaches, i.e. {\it TransH} \cite{DBLP:conf/aaai/WangZFC14}, {\it TransM}\cite{fan2014transition} and {\it TransE} \cite{Bordes2013a}, the proposed {\it PBE} approach still outperforms them with the improvements that {\bf NELL-50K}: \{{\it Accuracy}: $7.9\%\Uparrow$\},  {\bf WN-100K}: \{{\it Accuracy}: $5.6\%\Uparrow$\} and {\bf FB-500K}: \{{\it Accuracy}: $5.6\%\Uparrow$\}, as shown in Table 6.

%

\section{Conclusion}
\label{sec:con}
This paper proposed an elegant probabilistic model to tackle the problem of embedding beliefs which contain both structured knowledge and unstructured free texts, by firstly measuring the probability of a given belief $\langle h, r, t, m \rangle$. To efficiently learn the embeddings for each entity, relation, and word in mentions, we also adopted the negative sampling technique to transform the original model and display the algorithm based on stochastic gradient descent to search for the optimal solution. Extensive knowledge completion experiments, including {\it entity inference}, {\it relation prediction} and {\it triplet classification}, showed that our approach achieves significant improvement when tested with three large-scale repositories, compared with other state-of-the-art methods.


\section*{Acknowledgement}
The paper is dedicated to all the members of CSLT\footnote{\url{http://cslt.riit.tsinghua.edu.cn/}} and Proteus Group \footnote{\url{http://nlp.cs.nyu.edu/index.shtml}}. It was supported by National Program on Key Basic Research Project
(973 Program) under Grant 2013CB329304 and National Science Foundation of China (NSFC) under Grant No. 61373075, when the first author was a joint-supervision Ph.D. candidate of Tsinghua University and New York University.
\bibliographystyle{acl}
\bibliography{refs}

\begin{thebibliography}{}

\bibitem[\protect\citename{Bollacker \bgroup et al.\egroup
  }2007]{Bollacker2007}
Kurt Bollacker, Robert Cook, and Patrick Tufts.
\newblock 2007.
\newblock Freebase: A shared database of structured general human knowledge.
\newblock In {\em AAAI}, volume~7, pages 1962--1963.

\bibitem[\protect\citename{Bollacker \bgroup et al.\egroup
  }2008]{Bollacker2008}
Kurt Bollacker, Colin Evans, Praveen Paritosh, Tim Sturge, and Jamie Taylor.
\newblock 2008.
\newblock Freebase: a collaboratively created graph database for structuring
  human knowledge.
\newblock In {\em Proceedings of the 2008 ACM SIGMOD international conference
  on Management of data}, pages 1247--1250. ACM.

\bibitem[\protect\citename{Bordes \bgroup et al.\egroup }2011]{Bordes2011}
Antoine Bordes, Jason Weston, Ronan Collobert, Yoshua Bengio, et~al.
\newblock 2011.
\newblock Learning structured embeddings of knowledge bases.
\newblock In {\em AAAI}.

\bibitem[\protect\citename{Bordes \bgroup et al.\egroup }2013]{Bordes2013a}
Antoine Bordes, Nicolas Usunier, Alberto Garcia-Duran, Jason Weston, and Oksana
  Yakhnenko.
\newblock 2013.
\newblock Translating embeddings for modeling multi-relational data.
\newblock In {\em Advances in Neural Information Processing Systems}, pages
  2787--2795.

\bibitem[\protect\citename{Bordes \bgroup et al.\egroup }2014]{Bordes2014}
Antoine Bordes, Xavier Glorot, Jason Weston, and Yoshua Bengio.
\newblock 2014.
\newblock A semantic matching energy function for learning with
  multi-relational data.
\newblock {\em Machine Learning}, 94(2):233--259.

\bibitem[\protect\citename{Carlson \bgroup et al.\egroup }2010a]{Carlson2010}
Andrew Carlson, Justin Betteridge, Bryan Kisiel, Burr Settles, Estevam
  R.~Hruschka Jr., and Tom~M. Mitchell.
\newblock 2010a.
\newblock Toward an architecture for never-ending language learning.
\newblock In {\em Proceedings of the Twenty-Fourth Conference on Artificial
  Intelligence (AAAI 2010)}.

\bibitem[\protect\citename{Carlson \bgroup et al.\egroup }2010b]{carlson-aaai}
Andrew Carlson, Justin Betteridge, Bryan Kisiel, Burr Settles, Estevam
  R.~Hruschka Jr., and Tom~M. Mitchell.
\newblock 2010b.
\newblock Toward an architecture for never-ending language learning.
\newblock In {\em Proceedings of the Twenty-Fourth Conference on Artificial
  Intelligence (AAAI 2010)}.

\bibitem[\protect\citename{Fan \bgroup et al.\egroup
  }2014a]{fan-EtAl:2014:P14-1}
Miao Fan, Deli Zhao, Qiang Zhou, Zhiyuan Liu, Thomas~Fang Zheng, and Edward~Y.
  Chang.
\newblock 2014a.
\newblock Distant supervision for relation extraction with matrix completion.
\newblock In {\em Proceedings of the 52nd Annual Meeting of the Association for
  Computational Linguistics (Volume 1: Long Papers)}, pages 839--849,
  Baltimore, Maryland, June. Association for Computational Linguistics.

\bibitem[\protect\citename{Fan \bgroup et al.\egroup }2014b]{fan2014transition}
Miao Fan, Qiang Zhou, Emily Chang, and Thomas~Fang Zheng.
\newblock 2014b.
\newblock Transition-based knowledge graph embedding with relational mapping
  properties.
\newblock In {\em Proceedings of the 28th Pacific Asia Conference on Language,
  Information, and Computation}, pages 328--337.

\bibitem[\protect\citename{Fan \bgroup et al.\egroup }2015a]{fan2015jointly}
Miao Fan, Kai Cao, Yifan He, and Ralph Grishman.
\newblock 2015a.
\newblock Jointly embedding relations and mentions for knowledge population.
\newblock {\em arXiv preprint arXiv:1504.01683}.

\bibitem[\protect\citename{Fan \bgroup et al.\egroup }2015b]{fan2015learning}
Miao Fan, Qiang Zhou, and Thomas~Fang Zheng.
\newblock 2015b.
\newblock Learning embedding representations for knowledge inference on
  imperfect and incomplete repositories.
\newblock {\em arXiv preprint arXiv:1503.08155}.

\bibitem[\protect\citename{Grishman}1997]{Grishman:1997:IET:645856.669801}
Ralph Grishman.
\newblock 1997.
\newblock Information extraction: Techniques and challenges.
\newblock In {\em International Summer School on Information Extraction: A
  Multidisciplinary Approach to an Emerging Information Technology}, SCIE '97,
  pages 10--27, London, UK, UK. Springer-Verlag.

\bibitem[\protect\citename{Jenatton \bgroup et al.\egroup }2012]{Jenatton2012}
Rodolphe Jenatton, Nicolas Le~Roux, Antoine Bordes, Guillaume Obozinski, et~al.
\newblock 2012.
\newblock A latent factor model for highly multi-relational data.
\newblock In {\em NIPS}, pages 3176--3184.

\bibitem[\protect\citename{Mikolov \bgroup et al.\egroup
  }2013]{mikolov2013distributed}
Tomas Mikolov, Ilya Sutskever, Kai Chen, Greg~S Corrado, and Jeff Dean.
\newblock 2013.
\newblock Distributed representations of words and phrases and their
  compositionality.
\newblock In C.J.C. Burges, L.~Bottou, M.~Welling, Z.~Ghahramani, and K.Q.
  Weinberger, editors, {\em Advances in Neural Information Processing Systems
  26}, pages 3111--3119.

\bibitem[\protect\citename{Miller}1995]{Miller1995}
George~A. Miller.
\newblock 1995.
\newblock Wordnet: a lexical database for english.
\newblock {\em Communications of the ACM}, 38(11):39--41.

\bibitem[\protect\citename{Mintz \bgroup et al.\egroup }2009]{mintz2009distant}
Mike Mintz, Steven Bills, Rion Snow, and Dan Jurafsky.
\newblock 2009.
\newblock Distant supervision for relation extraction without labeled data.
\newblock In {\em Proceedings of the Joint Conference of the 47th Annual
  Meeting of the ACL and the 4th International Joint Conference on Natural
  Language Processing of the AFNLP: Volume 2-Volume 2}, pages 1003--1011.
  Association for Computational Linguistics.

\bibitem[\protect\citename{Mitchell \bgroup et al.\egroup }2015]{NELL-aaai15}
T.~Mitchell, W.~Cohen, E.~Hruschka, P.~Talukdar, J.~Betteridge, A.~Carlson,
  B.~Dalvi, M.~Gardner, B.~Kisiel, J.~Krishnamurthy, N.~Lao, K.~Mazaitis,
  T.~Mohamed, N.~Nakashole, E.~Platanios, A.~Ritter, M.~Samadi, B.~Settles,
  R.~Wang, D.~Wijaya, A.~Gupta, X.~Chen, A.~Saparov, M.~Greaves, and
  J.~Welling.
\newblock 2015.
\newblock Never-ending learning.
\newblock In {\em Proceedings of the Twenty-Ninth AAAI Conference on Artificial
  Intelligence (AAAI-15)}.

\bibitem[\protect\citename{Sarawagi}2008]{sarawagi2008information}
Sunita Sarawagi.
\newblock 2008.
\newblock Information extraction.
\newblock {\em Foundations and trends in databases}, 1(3):261--377.

\bibitem[\protect\citename{Socher \bgroup et al.\egroup }2013]{Socher2013}
Richard Socher, Danqi Chen, Christopher~D Manning, and Andrew Ng.
\newblock 2013.
\newblock Reasoning with neural tensor networks for knowledge base completion.
\newblock In {\em Advances in Neural Information Processing Systems}, pages
  926--934.

\bibitem[\protect\citename{Sutskever \bgroup et al.\egroup
  }2009]{Sutskever2009}
Ilya Sutskever, Ruslan Salakhutdinov, and Joshua~B Tenenbaum.
\newblock 2009.
\newblock Modelling relational data using bayesian clustered tensor
  factorization.
\newblock In {\em NIPS}, pages 1821--1828.

\bibitem[\protect\citename{Wang \bgroup et al.\egroup }2014a]{D14-1167}
Zhen Wang, Jianwen Zhang, Jianlin Feng, and Zheng Chen.
\newblock 2014a.
\newblock Knowledge graph and text jointly embedding.
\newblock In {\em Proceedings of the 2014 Conference on Empirical Methods in
  Natural Language Processing (EMNLP)}, pages 1591--1601. Association for
  Computational Linguistics.

\bibitem[\protect\citename{Wang \bgroup et al.\egroup
  }2014b]{DBLP:conf/aaai/WangZFC14}
Zhen Wang, Jianwen Zhang, Jianlin Feng, and Zheng Chen.
\newblock 2014b.
\newblock Knowledge graph embedding by translating on hyperplanes.
\newblock In {\em Proceedings of the Twenty-Eighth {AAAI} Conference on
  Artificial Intelligence, July 27 -31, 2014, Qu{\'{e}}bec City, Qu{\'{e}}bec,
  Canada.}, pages 1112--1119.

\bibitem[\protect\citename{West \bgroup et al.\egroup }2014]{42024}
Robert West, Evgeniy Gabrilovich, Kevin Murphy, Shaohua Sun, Rahul Gupta, and
  Dekang Lin.
\newblock 2014.
\newblock Knowledge base completion via search-based question answering.
\newblock In {\em WWW}.

\bibitem[\protect\citename{Weston \bgroup et al.\egroup
  }2013]{weston-EtAl:2013:EMNLP}
Jason Weston, Antoine Bordes, Oksana Yakhnenko, and Nicolas Usunier.
\newblock 2013.
\newblock Connecting language and knowledge bases with embedding models for
  relation extraction.
\newblock In {\em Proceedings of the 2013 Conference on Empirical Methods in
  Natural Language Processing}, pages 1366--1371, Seattle, Washington, USA,
  October. Association for Computational Linguistics.

\end{thebibliography}

\end{document}